\documentclass[letterpaper,10pt,conference]{ieeeconf}

\IEEEoverridecommandlockouts
\usepackage{graphicx}

\usepackage{amsmath}
\usepackage{amssymb}
\usepackage{bm}

\usepackage{booktabs}
\usepackage{array}
\usepackage{capt-of}
\usepackage{cite}
\usepackage{float}
\usepackage{stfloats}
\usepackage{flushend}
\usepackage{microtype}
\usepackage{placeins}
\usepackage{tikz}
\usetikzlibrary{arrows.meta,backgrounds,calc,fit,positioning}


\graphicspath{{figures/}}

\title{
HuMemSLAM: Efficient Human-Inspired Semantic Place Recognition
for Robust Visual SLAM
}

\author{Mayowa Adebambo, Sebastian Donnelly, Armand Amaritei, Andrew Bradley, Alexander Rast%
\thanks{School of Engineering, Computing \& Mathematics, Oxford Brookes
University, Oxford, UK.}%
\thanks{Autonomous Driving and Intelligent Transport Group, Oxford Brookes
University, Oxford, UK.} \thanks{This work utilised the experimental vehicle
developed through the GREEN-LOG project (Horizon Europe Grant Agreement No.
101069892) \cite{11654478}.}}

\begin{document}

\maketitle
\pagestyle{empty}
\raggedbottom

% !TeX root = ../main.tex

\begin{abstract}

Autonomous systems require reliable place recognition for efficient and
effective simultaneous localisation and mapping (SLAM).\ Traditional geometric
visual SLAM approaches rely on low-level features and geometric consistency, but
remain vulnerable to perceptual aliasing,\ where different places appear
similar, and perceptual variation, where the same place appears different.\
Although semantic SLAM and modern learned visual place recognition (VPR) methods
improve robustness under challenging perceptual conditions,\ real-time
deployment requires both high retrieval accuracy and low latency.\ Inspired by
human memory and perception, we propose HuMem-VPR, which exploits the
bidirectional relationship between bottom-up perceptual evidence and top-down
contextual reasoning to achieve high-level place understanding.\ We further
introduce HuMemSLAM, the integration of HuMem-VPR with ORB-SLAM3. HuMem-VPR
achieved the highest aggregate retrieval accuracy on the real-image benchmark,
competitive accuracy on the CARLA benchmark, and approximately two to three
times lower latency than the evaluated state-of-the-art VPR methods.\ Across the
evaluated dataset families and online experiments,\ HuMemSLAM substantially
improved integrated Recall~@1 over ORB-SLAM3's native retrieval while reducing
the proposals submitted to its geometric backend.

\end{abstract}

% !TeX root = ../main.tex

\section{Introduction}
\label{sec:introduction}

Simultaneous Localisation and Mapping (SLAM) estimates an autonomous agent's
pose while constructing or maintaining a representation of its surroundings.
Place recognition, determining whether the current location was previously
visited, supports loop closure, relocalisation, drift correction, and map fusion
\cite{campos2021orbslam3}. As these operations occur online, place recognition
must be both accurate and sufficiently fast for real-time deployment.

Geometric visual SLAM typically relies on low-level appearance and geometric
consistency. ORB-SLAM, for example, uses ORB features and a Bag-of-Words (BoW)
representation for retrieval before geometric verification
\cite{campos2021orbslam3,galvezlopez2012dbow2}. This creates two long-term
challenges: perceptual aliasing, where distinct but repetitive locations appear
similar, and perceptual variation, where illumination, season, weather, blur, or
viewpoint makes the same place appear different. Severe change can degrade image
texture and local correspondences, weakening both retrieval and geometric
matching \cite{lowry2016vprsurvey,masone2021deepvpr}.

Learned visual place recognition (VPR) and semantic SLAM improve robustness
through higher-level representations that can remain discriminative when
low-level appearance changes
\cite{berton2023eigenplaces,keetha2024anyloc,izquierdo2024salad,berton2025megaloc}.
Yet online localisation requires not only accuracy and robustness to perceptual
variation and aliasing, but also sufficiently low latency
\cite{labbe2019rtabmap}. In autonomous driving, perception and localisation
must satisfy tight timing constraints to support safe operation, particularly
at high speeds \cite{liu2019edgeautonomous}. Although recent VPR work addresses
computational efficiency \cite{alibey2023mixvpr}, latency remains an important
deployment consideration. This motivates our central question:
\emph{can human-inspired structured high-level semantic reasoning provide
accurate and robust place hypotheses while preserving the low latency required
by online visual SLAM?}

\begin{figure}[!ht]
    \centering
    \includegraphics[width=0.78\columnwidth]{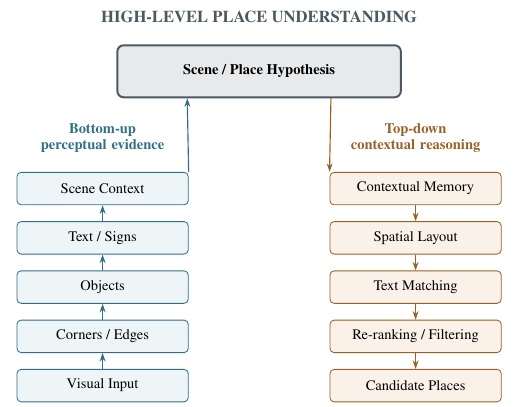}
    \caption{Conceptual motivation for HuMemSLAM: bottom-up perceptual evidence and top-down contextual reasoning for high-level place understanding.}
    \label{fig:place-understanding}
\end{figure}

Research in human visual cognition suggests that recognition combines bottom-up evidence with top-down context \cite{oliva2006gist,bar2006topdown}, as illustrated in Fig.~\ref{fig:place-understanding}. We therefore propose HuMemSLAM, a human-inspired semantic place-recognition framework integrated with geometric visual SLAM. \emph{HuMem-VPR} denotes its standalone semantic retrieval subsystem; \emph{HuMemSLAM} denotes HuMem-VPR integrated with ORB-SLAM3. HuMem-VPR ranks historical keyframes using high-level semantics, while ORB-SLAM3 verifies the resulting hypotheses and performs relocalisation, loop correction, or map fusion.

The main contributions of this work are:

\begin{enumerate}
    \item HuMem-VPR, a human-inspired place recognition algorithm that corroborates holistic hypotheses of the scene representation with object-spatial and object-grounded text information.

    \item HuMemSLAM, the integration of HuMem-VPR with ORB-SLAM3 as a semantic retrieval layer over its existing map.
\end{enumerate}

% !TeX root = ../main.tex

\section{Related Work}
\label{sec:related-work}

\subsection{Visual Place Recognition under Perceptual Change}

Visual place recognition (VPR) identifies previously observed locations despite viewpoint, illumination, weather, and seasonal change. Such variation and aliasing between visually similar but distinct places remain central challenges for long-term localisation \cite{lowry2016vprsurvey,masone2021deepvpr}. Modern VPR has increasingly shifted from handcrafted features and visual vocabularies towards learned global representations.

NetVLAD established end-to-end learned descriptor aggregation for large-scale place recognition \cite{arandjelovic2016netvlad}. EigenPlaces trains viewpoint-robust descriptors \cite{berton2023eigenplaces}, while AnyLoc uses self-supervised foundation-model representations for cross-domain generalisation without VPR-specific retraining \cite{keetha2024anyloc}. SALAD combines DINOv2 features with optimal-transport aggregation \cite{izquierdo2024salad}; MegaLoc targets transfer across VPR, localisation, and landmark retrieval \cite{berton2025megaloc}. These approaches demonstrate the strength of learned
global representations, but also illustrate that contemporary
VPR systems occupy different trade-offs between retrieval
accuracy, generalisation, robustness, and computational cost.

SVS-VPR uses aggregated semantic classes for coarse filtering, followed by learned local correspondences constrained by semantics and spatial consistency \cite{arshad2024svsvpr}. TextPlace uses recognised text and its spatio-temporal relationships as distinctive landmarks \cite{hong2019textplace}. HuMem-VPR instead begins with a holistic scene representation and corroborates it through object-instance, spatial-layout, and object-grounded textual reasoning.

\subsection{Place Recognition within Visual SLAM}

Within visual SLAM, retrieved places become hypotheses for geometric verification, loop closure, relocalisation, and map fusion. Classical systems use efficient BoW retrieval: DBoW2 combines binary visual words with geometric verification \cite{galvezlopez2012dbow2}; iBoW-LCD learns its vocabulary incrementally \cite{garciafidalgo2018ibowlcd}; and RTAB-Map couples appearance-based loop closure with memory management \cite{labbe2019rtabmap}.

ORB-SLAM3 uses retrieved keyframes for relocalisation, loop closure, and map merging while retaining geometric consistency checks \cite{campos2021orbslam3}. SLGD-Loop adds global semantic similarity and salient local features in a coarse-to-fine detector for long-term appearance change \cite{arshad2024slgdloop}. These systems distinguish plausible retrieval from a valid geometric constraint; embedded VPR should therefore be assessed by both retrieval accuracy and the hypotheses passed to verification.

HuMem-VPR ranks historical ORB-SLAM3 keyframes, while the geometric backend establishes physical consistency and applies SLAM updates. We consequently evaluate it both standalone and through its effect on the downstream geometric pipeline.

\subsection{Biologically and Cognitively Inspired Place Recognition}

Biological navigation has motivated alternative localisation representations. RatSLAM combines self-motion and landmarks through a continuous-attractor model of rodent hippocampal navigation \cite{milford2004ratslam}. Multi-scale extensions use parallel maps \cite{chen2015multiscale} or adaptive scale selection with coarse-to-fine recognition \cite{fan2017adaptive}. LPMP instead combines learned landmark identities (``what'') with spatial arrangement (``where'') \cite{colomer2022lpmp}.

HuMemSLAM draws from visual cognition rather than reproducing a neural circuit. Global scene structure can rapidly convey scene gist \cite{oliva2006gist}, while visual-recognition models emphasise interaction between perceptual evidence and top-down contextual predictions \cite{bar2006topdown}. HuMemSLAM establishes a global scene hypothesis, then adds object-spatial and grounded-text evidence. Unlike spatial-cell models, this semantic representation is a retrieval layer over the ORB-SLAM3 map; geometry remains the final authority.

% !TeX root = ../main.tex

\section{HuMemSLAM Framework}
\label{sec:methodology}

\subsection{System Overview}
\label{subsec:system_overview}

HuMemSLAM integrates the standalone HuMem-VPR semantic retrieval subsystem with ORB-SLAM3 \cite{campos2021orbslam3}. HuMem-VPR indexes ORB-SLAM3 keyframes for retrieval under appearance change, while the backend retains pose estimation and geometric verification.

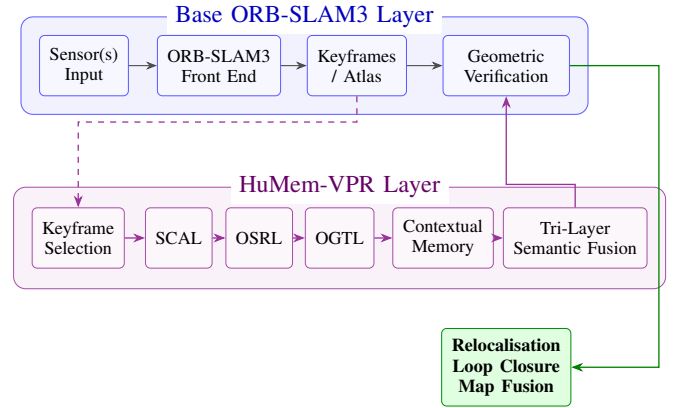
\begin{figure}[H]
    \centering
    \resizebox{\columnwidth}{!}{%
    \begin{tikzpicture}[
        font=\small,
        >=Stealth,
        block/.style={draw, rounded corners=2.5pt, minimum height=1.05cm,
            align=center, inner xsep=5pt, fill=white},
        baseblock/.style={block, draw=blue!65, fill=blue!3},
        contextblock/.style={block, draw=violet!70, fill=violet!3},
        outputblock/.style={block, draw=green!55!black, fill=green!12,
            minimum width=1.65cm, minimum height=1.8cm, font=\small\bfseries},
        flow/.style={->, line width=0.55pt, draw=black!70},
        baseflow/.style={->, line width=0.65pt, draw=green!50!black},
        contextflow/.style={->, line width=0.65pt, draw=violet!80},
        feedback/.style={->, dashed, line width=0.65pt, draw=violet!80}
    ]
        % Compact ORB-SLAM3 path
        \node[baseblock, minimum width=1.45cm] (sensors) at (0,5.35)
            {Sensor(s)\\Input};
        \node[baseblock, minimum width=2.05cm] (frontend) at (2.35,5.35)
            {ORB-SLAM3\\Front End};
        \node[baseblock, minimum width=1.70cm] (atlas) at (4.75,5.35)
            {Keyframes\\/ Atlas};
        \node[baseblock, minimum width=2.20cm] (verification) at (7.35,5.35)
            {Geometric\\Verification};

        \draw[flow] (sensors) -- (frontend);
        \draw[flow] (frontend) -- (atlas);
        \draw[flow] (atlas) -- (verification);

        % Sequential human-inspired semantic reasoning
        \node[contextblock, minimum width=1.40cm, minimum height=1.05cm] (selection) at (-0.10,2.35)
            {Keyframe\\Selection};
        \node[contextblock, minimum width=1.10cm, minimum height=1.05cm] (scal) at (1.65,2.35)
            {SCAL};
        \node[contextblock, minimum width=1.10cm, minimum height=1.05cm] (osrl) at (3.05,2.35)
            {OSRL};
        \node[contextblock, minimum width=1.10cm, minimum height=1.05cm] (ogtl) at (4.45,2.35)
            {OGTL};
        \node[contextblock, minimum width=1.75cm, minimum height=1.05cm] (memory) at (6.25,2.35)
            {Contextual\\Memory};
        \node[contextblock, minimum width=1.90cm, minimum height=1.05cm] (fusion) at (8.55,2.35)
            {Tri-Layer\\Semantic Fusion};

        \draw[feedback] (atlas.south) -- ++(0,-0.45) -| (selection.north);
        \draw[contextflow] (selection) -- (scal);
        \draw[contextflow] (scal) -- (osrl);
        \draw[contextflow] (osrl) -- (ogtl);
        \draw[contextflow] (ogtl) -- (memory);
        \draw[contextflow] (memory) -- (fusion);
        \draw[contextflow] (fusion.north) -- ++(0,0.40) -| (verification.south);

        % Verified SLAM actions
        \node[outputblock, minimum height=1.35cm] (output) at (7.35,0.10)
            {Relocalisation\\Loop Closure\\Map Fusion};
        \draw[baseflow] (verification.east) -- ++(1.55,0) |- (output.east);

        % Layer containers and titles
        \begin{scope}[on background layer]
            \node[draw=blue!65, fill=blue!5, rounded corners=6pt,
                fit=(sensors)(frontend)(atlas)(verification),
                inner xsep=9pt, inner ysep=9pt] (basebox) {};
            \node[draw=violet!70, fill=violet!5, rounded corners=6pt,
                fit=(selection)(scal)(osrl)(ogtl)(memory)(fusion),
                inner xsep=9pt, inner ysep=10pt] (contextbox) {};
        \end{scope}
        \node[font=\large, text=blue!75!black, fill=white, inner xsep=4pt]
            at (basebox.north) {Base ORB-SLAM3 Layer};
        \node[font=\large, text=violet!80!black, fill=white, inner xsep=4pt]
            at (contextbox.north) {HuMem-VPR Layer};
    \end{tikzpicture}%
    }
    \caption{HuMemSLAM architecture. HuMem-VPR sequentially applies scene-context, object-spatial, and object-grounded text reasoning; ORB-SLAM3 retains geometric verification and map updates.}
    \label{fig:humanslam-architecture}
\end{figure}

For query keyframe $Q_t$, the semantic memory and bounded similarity are
\begin{equation}
    \mathcal{D}_{t-1}=\{K_1,\ldots,K_{t-1}\},
    \qquad \Lambda(Q_t,K_i)\in[0,1].
\end{equation}
To avoid trivial temporal neighbours, the eligible set is
$\mathcal{C}_t=\{K_i\in\mathcal{D}_{t-1}:|f_t-f_i|\geq100\}$, where $f$
denotes the source-frame identifier. Candidates are ordered by
\begin{equation}
    \pi_t=\operatorname{argsort}^{\downarrow}_{K_i\in\mathcal{C}_t}
    \Lambda(Q_t,K_i).
\end{equation}
HuMem-VPR proposes \emph{where} the camera may have been; ORB-SLAM3 tests geometric consistency and may trigger relocalisation, loop closure, map fusion, or pose correction (Fig.~\ref{fig:humanslam-architecture}).

\FloatBarrier

\subsection{Semantic Keyframe Representation}
\label{subsec:keyframe_representation}

Each semantic keyframe is represented as

\begin{equation}
    K_i =
    \left(
    i,f_i,t_i,m_i,\mathbf{T}_i,S_i,\mathcal{O}_i
    \right),
\end{equation}

where $i$ is the ORB-SLAM3 keyframe identifier, $f_i$ is the source-frame
identifier, $t_i$ is the timestamp, $m_i$ is the Atlas map identifier,
$\mathbf{T}_i \in SE(3)$ is the estimated camera pose, $S_i$ is the
scene-context representation, and $\mathcal{O}_i$ is the set of detected
stable objects. Each object record stores its semantic class, detector
confidence, spatial attributes, and associated textual observations.

%The semantic representation preserves the original ORB-SLAM3 identifiers.
%HuMem-VPR therefore augments the existing geometric map with a semantic
%retrieval index rather than constructing an independent geometric map.

\subsection{Scene Context Attention Layer (SCAL)}
\label{subsec:sam}

The Scene Context Attention Layer (SCAL) provides the initial global place hypothesis. For an image
$I_i$, a 512-dimensional EigenPlaces descriptor \cite{berton2023eigenplaces}
$\mathbf{e}_i \in \mathbb{R}^{512}$ is extracted. Its normalisation and
non-negative cosine similarity are
\begin{equation}
    \hat{\mathbf e}_i=\frac{\mathbf e_i}{\|\mathbf e_i\|_2},
    \qquad
    s_e(Q,C)=\operatorname{clip}
    (\hat{\mathbf e}_Q^\top\hat{\mathbf e}_C,0,1).
\end{equation}

Places365 scene labels provide contextual modulation \cite{zhou2018places}. Let
$C_{\mathrm{cat}}(Q,C)\in[0,1]$ denote the compatibility between the
scene categories of the query and candidate. With $\lambda_c=0.10$, the
modulated similarity is

\begin{equation}
    \tilde{s}_e(Q,C)
    =
    s_e(Q,C)
    \frac{
        1+\lambda_c C_{\mathrm{cat}}(Q,C)
    }{
        1+\lambda_c
    }.
\end{equation}

Short-term context combines the current and up to two preceding
observations using $\boldsymbol\alpha=(0.5,0.3,0.2)$. The resulting score is

\begin{equation}
    S_s(Q,C)
    =
    \sum_{j=0}^{k-1}
    \alpha_j
    \tilde{s}_e
    \left(
    Q_{t-j},C_{i-j}
    \right),
    \qquad
    \sum_j \alpha_j = 1,
\end{equation}

where $k\leq3$ depends on the available temporal history.

SCAL passes only its top $K=25$ candidates to object and textual comparison.

\subsection{Object-Spatial Reasoning Layer (OSRL)}
\label{subsec:osrl}

The Object-Spatial Reasoning Layer (OSRL) evaluates whether candidate objects have spatial arrangements consistent with the query. A YOLOv26 segmentation model \cite{jocher2026yolo26} supplies
the semantic class, confidence, and image region for each detected object;
HuMem-VPR derives the normalised centroid and area used for spatial
reasoning from these regions. The YOLOv26 model is fine-tuned on the Mapillary Vistas dataset \cite{neuhold2017mapillary} and classes are restricted to predefined static environmental structures, including buildings, fences, street lights and billboards. 

For a query object $q$ and candidate object $c$, let $(x,y)$ denote the
normalised object centroid and let $a$ denote its normalised image area. Their
squared spatial-layout distance is

\begin{equation}
    d_m^2(q,c)
    =
        (x_q-x_c)^2
        +
        (y_q-y_c)^2
        +
        \lambda_a(a_q-a_c)^2.
\end{equation}

With $\lambda_a=1$, a Gaussian spatial kernel, same-class gate, and
detector confidences $p_q,p_c$ define the pairwise score
\begin{align}
    G(q,c)&=\exp\!\left[-\frac{d_m^2(q,c)}{2\sigma_m^2}\right],
        \qquad \sigma_m=0.25,\\
    \delta_c(q,c)&=\mathbf 1[c_q=c_c],\\
    M_o(q,c)&=\delta_c(q,c)\sqrt{p_qp_c}\,G(q,c).
\end{align}

For each semantic class $r$, HuMem-VPR constructs a pairwise similarity
matrix and solves a maximum-weight one-to-one bipartite assignment using
the Hungarian algorithm:

\begin{equation}
    \mathcal{A}_r^\star
    =
    \arg\max_{\mathcal{A}_r}
    \sum_{(u,v)\in\mathcal{A}_r}
    M^{(r)}_{uv}.
\end{equation}

The final object-spatial similarity is query-normalised:

\begin{equation}
    S_o(Q,C)
    =
    \frac{
        \displaystyle
        \sum_r
        \omega_r
        \sum_{(u,v)\in\mathcal{A}_r^\star}
        M^{(r)}_{uv}
    }{
        \displaystyle
        \sum_{q\in\mathcal{O}_Q}
        \omega_{c_q}
    },
\end{equation}

where $\omega_r$ denotes the configured importance of semantic class $r$.
The query-dependent denominator penalises missing candidate evidence and sparse, easily matched objects.

\subsection{Object-Grounded Textual Landmark Layer (OGTL)}
\label{subsec:tll}

The Object-Grounded Textual Landmark Layer (OGTL) treats text as a
place-specific landmark only when it is grounded in a compatible semantic
object. PP-OCRv5 Mobile \cite{cui2026ppocrv5} performs text detection and English text
recognition within the detected object regions, and each resulting textual
observation is stored in the corresponding object record. OCR observations
are compared only when their supporting objects share the same semantic
class and have sufficient spatial compatibility:

\begin{equation}
    \Gamma_T(q,c)
    =
    \mathbf{1}[c_q=c_c]\,
    \mathbf{1}[G(q,c)\geq\tau_g],
    \qquad
    \tau_g=0.6.
\end{equation}

For strings $a,b$, HuMem-VPR takes the strongest of normalised
Levenshtein, token Jaccard, and compact substring similarities:
\begin{align}
 L(a,b)&=1-\frac{D_{\mathrm{lev}}(a,b)}{\max(|a|,|b|,1)},\\
 J(a,b)&=\frac{|\operatorname{tok}(a)\cap\operatorname{tok}(b)|}
 {|\operatorname{tok}(a)\cup\operatorname{tok}(b)|},\\
 S_{\mathrm{str}}(a,b)&=\max\{L(a,b),J(a,b),C(a,b)\},\\
 M_T(u,v)&=S_{\mathrm{str}}(u,v)p_up_v.
\end{align}

HuMem-VPR additionally assigns each textual observation a distinctiveness
weight $D_T(u)\in[0,1]$, giving greater importance to longer, uncommon, and
digit-bearing strings while reducing the influence of generic words.

For a text-bearing query object $q$ and candidate object $c$, the
object-grounded text score is

\begin{equation}
    S_T(q,c)
    =
    \frac{
        \displaystyle
        \sum_{u\in\mathcal{T}_q}
        D_T(u)
        \max_{v\in\mathcal{T}_c}
        M_T(u,v)
    }{
        \displaystyle
        \sum_{u\in\mathcal{T}_q}
        D_T(u)
    }.
\end{equation}

The keyframe-level textual similarity $S_t$ is modulated by
$E_t=\sqrt{D_QD_C}$, where $D_Q,D_C$ are the mean textual distinctiveness
of the query and candidate. Distinctive landmarks therefore contribute
more strongly than weak or generic OCR observations.

\subsection{Tri-Layer Callosal Semantic Fusion}
\label{subsec:fusion}

The scene, object-spatial, and textual signals are combined using a
scene-primary semantic fusion strategy. Scene recognition provides the
initial place hypothesis, while object and textual information act as
bounded corroborating evidence.

Object and textual support are first combined using $g_o=0.15$ and
$g_t=0.25$:

\begin{equation}
    B(Q,C)
    =
    \operatorname{clip}
    \left(
        g_oS_o
        +
        g_tE_tS_t,
        0,1
    \right).
\end{equation}

The final raw HuMem-VPR similarity is

\begin{equation}
    \Lambda_{\mathrm{raw}}(Q,C)
    =
    S_s
    +
    (1-S_s)B(Q,C),
    \label{eq:humanslam_fusion}
\end{equation}

Equation~\eqref{eq:humanslam_fusion} gives the fusion an intuitive
interpretation: semantic object and textual evidence reduce the remaining
uncertainty in the scene-level hypothesis.

\subsection{Candidate Selection and Geometric Verification}
\label{subsec:candidate_verification}

Historical candidates are ranked using the fused HuMem-VPR similarity
and are submitted for geometric verification when

\begin{equation}
    \Lambda_{\mathrm{raw}}(Q,C) > \tau_s,
    \qquad
    \tau_s=0.70.
\end{equation}

Up to five ranked semantic candidates are returned to ORB-SLAM3. During
normal tracking, semantic retrieval is performed periodically, while
retrieval is activated immediately when ORB-SLAM3 enters the
\texttt{RECENTLY\_LOST} or \texttt{LOST} states. HuMem-VPR returns its semantic hypothesis to ORB-SLAM3, which then performs local-feature
correspondence and robust geometric verification before accepting a
relocalisation, loop closure, map fusion, or pose correction.

\subsection{Low-Latency Computational Design}
\label{subsec:Low-Lantency Design}
For $N$ global descriptors of dimension $d$, the initial SCAL retrieval has
complexity $O(Nd)$, while subsequent object-spatial and textual reasoning is
restricted to the top $K=25$ candidates. Likewise, the Hungarian assignment,
with worst-case complexity $O(n_r^3)$ for a semantic class containing $n_r$
objects, is applied only to small same-class object sets rather than the full
semantic memory.

Neural inference is accelerated using TensorRT to reduce the runtime cost of
the learned perception components. HuMem-VPR further operates asynchronously
with respect to ORB-SLAM3 tracking and processes selected keyframes rather than
every incoming frame. A bounded processing queue replaces stale pending work
with the newest query when full, prioritising current localisation information
and preventing semantic processing from blocking the tracking thread.
% !TeX root = ../main.tex

\section{Experimental Setup}
\label{sec:experimental-setup}
ORB-SLAM3's asynchronous modules can vary across runs because of thread
scheduling, resource contention, and CPU/GPU timing. We therefore repeat every
offline full SLAM and VPR comparison involving ORB-SLAM3 ten times and report
arithmetic means; tracking failures and incomplete trajectories are reported
separately.

\subsection{Datasets and Perceptual Conditions}
\label{subsec:datasets}

Table~\ref{tab:datasets-conditions} summarises the datasets and evaluated conditions.

\begin{figure*}[!t]
    \centering
    \includegraphics[width=\textwidth]{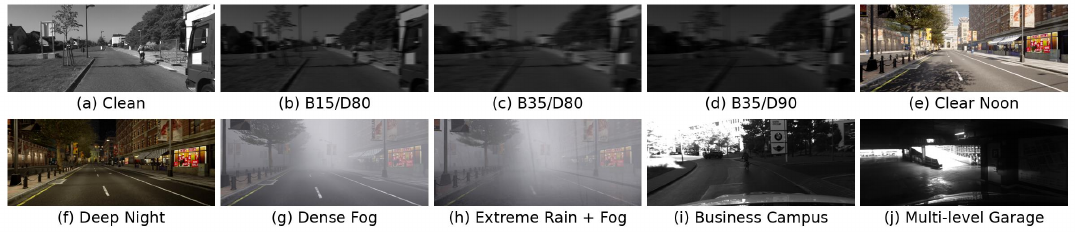}
    \caption{Evaluation datasets and perceptual conditions.}
    \label{fig:evaluation-conditions}
\end{figure*}

The three dataset families provide complementary perceptual challenges as shown in Fig.~\ref{fig:evaluation-conditions}. KITTI evaluates controlled image degradation on real driving imagery: sequence 06 corrupts the return traversal, whereas sequences 00 and 05 perturb only ground-truth-identified revisit regions. A condition denoted \(B\ell/Dd\) applies a line-motion kernel of length \(\ell\) pixels followed by exposure reduction retaining \((100-d)\%\) of linear-light intensity. CARLA compares repeated traversals of the same route under controlled simulated weather and illumination changes. 4Seasons evaluates naturally occurring seasonal appearance variation and perceptual aliasing in structurally repetitive environments.

\begin{table}[H]
    \caption{Datasets and perceptual conditions used in the evaluation.}
    \label{tab:datasets-conditions}
    \centering
    \footnotesize
    \setlength{\tabcolsep}{2.2pt}
    \renewcommand{\arraystretch}{1.03}
    \begin{tabular}{@{}>{\raggedright\arraybackslash}p{0.24\columnwidth}>{\raggedright\arraybackslash}p{0.27\columnwidth}>{\raggedright\arraybackslash}p{0.43\columnwidth}@{}}
        \toprule
        Dataset / route & Main challenge & Evaluated conditions \\
        \midrule
        \textbf{KITTI 00/05/06} \cite{geiger2012kitti}
            & Controlled blur and illumination/exposure degradation on real stereo road imagery with ground-truth poses
            & Seq. 06: Clean, B15, D50, B15/D50, B15/D80, B35/D80, B35/D90; Seq. 00/05: Clean, B15/D80 \\
        \addlinespace[3pt]
        \textbf{CARLA}\newline Town10HD Opt \cite{dosovitskiy2017carla}
            & Controlled weather and illumination change over repeated traversals
            & Clear Noon (reference), Deep Night, Dense Fog Overcast, Extreme Rain + Fog, Extreme Sunset Glare \\
        \addlinespace[3pt]
        \textbf{4Seasons}\newline Business Campus \cite{wenzel2020fourseasons}
            & Natural seasonal appearance variation
            & Fall, Winter \\
        \addlinespace[3pt]
        \textbf{4Seasons}\newline Multi-level Garage
            & Perceptual aliasing across structurally similar locations and floors
            & Dec., Feb., May traversals \\
        \bottomrule
    \end{tabular}
\end{table}

\subsection{Query, Database and Comparison Protocol}
\label{subsec:vpr_protocol}

The fixed VPR benchmark uses deterministic manifests shared by HuMem-VPR, native ORB-SLAM3 ORB/DBoW2, SALAD, and MegaLoc. A database frame \(c\) is a valid historical match for query \(q\) if:
\begin{equation}
  \begin{aligned}
    q-c &\geq 100, &
    \lVert\mathbf{t}_q-\mathbf{t}_c\rVert_2 &\leq 5\,\mathrm{m}, \\
    \Delta R(q,c) &\leq 30^{\circ}. &&
  \end{aligned}
\end{equation}
where \(\Delta R\) is the geodesic orientation difference computed from rotation matrices for KITTI/CARLA and unit quaternions for 4Seasons. Only queries having at least one valid database match are eligible; at most 100 eligible queries are selected deterministically at approximately uniform index intervals. At query time, every method is restricted to database frames satisfying the 100-frame exclusion.

\subsection{HuMem-VPR Ablation Configurations}
\label{subsec:ablations}

End-to-end layer ablations use four implemented configurations: SCAL, SCAL+OSRL,
SCAL+OGTL, and Full (SCAL+OSRL+OGTL). In SCAL+OGTL, object similarity is
disabled; however, YOLO remains active solely to provide object regions for OCR.
% TODO: verify whether the paper reports the legacy tri-layer KITTI ablation, the current scene-support Business Campus ablation, or both; their fusion rules must not be pooled [NEEDS VERIFICATION].

\subsection{VPR Metrics and Latency Evaluation}
\label{subsec:vpr_metrics}

Let \(r_q\) be the rank of the first valid historical match for query \(q\), or infinity if none is returned. For \(N\) eligible queries,
\begin{equation}
  \begin{aligned}
    \mathrm{Recall}\,@K &= \frac{1}{N}\sum_q\mathbb{I}[r_q\leq K], \\
    \mathrm{MRR} &= \frac{1}{N}\sum_q\frac{1}{r_q}.
  \end{aligned}
\end{equation}
where \(1/\infty=0\). We report Recall~@1, Recall~@5, and MRR. Latency uses 30 deterministically spaced queries (or all if fewer), ten neural-model warm-ups, and synchronised CUDA timing; image decoding and database construction are excluded. HuMem-VPR timing spans scene description, segmentation, OCR, semantic fusion, database comparison, and top-five ranking. ORB BoW spans grayscale conversion, native ORB extraction, vocabulary transformation, DBoW2 comparison, and ranking. Official PyTorch SALAD and MegaLoc implementations \cite{izquierdo2024salad,berton2025megaloc} include preprocessing, inference, normalisation, cosine search, and ranking. We report mean, median, and 95th-percentile latency.

\subsection{Integrated SLAM Evaluation}
\label{subsec:slam_eval}
At SLAM level, proposals use the same 5 m, 30\(^{\circ}\), and 100-frame match
rule for the offline runs; online matches must be within 5 m and at least 30
seconds older than the query. Recall~@1, Recall~@5, and MRR include logged
queries with a qualifying historical keyframe. Geometry attempts, accepted
seeds, loop corrections, and latencies are reported independently. A
relocalisation episode starts in \texttt{RECENTLY\_LOST} or \texttt{LOST};
recovery is a return to \texttt{OK}.

For 4Seasons garage runs, cross-floor aliasing requires at most 5 m horizontal
separation and at least 2.4 m height difference; both denied and accepted
proposals are retained. Secondary measures are absolute pose error (APE),
trajectory completeness, and map count. Source-frame-associated trajectories are
evaluated per map with rigid alignment using \texttt{evo\_ape}
\cite{grupp2017evo}; global APE is reported only for single-map trajectories.

\subsection{Computational and Hardware Setup}
\label{subsec:hardware}

Recorded runs were executed on an Intel Core Ultra-series 185H host with
approximately 30.7 GiB RAM and an NVIDIA GeForce RTX 4070 Laptop GPU with 8188
MiB memory. The recorded environment is Ubuntu 22.04.5 LTS, ROS~2 Humble, CUDA
12.9, TensorRT 11.0.0.114, and PyTorch 2.8.0+cu129. 
% TODO: add compiler, OpenCV build, ORB-SLAM3 build flags, and thread-affinity policy if required [NEEDS VERIFICATION].

% Pending: \subsection{Real-Time Vehicle Evaluation}\label{subsec:vehicle_eval}

\subsection{Real-World Online Experiment}
\label{subsec:eval}

\begin{figure}[H]
    \centering
    \includegraphics[height=0.75422\columnwidth,keepaspectratio]{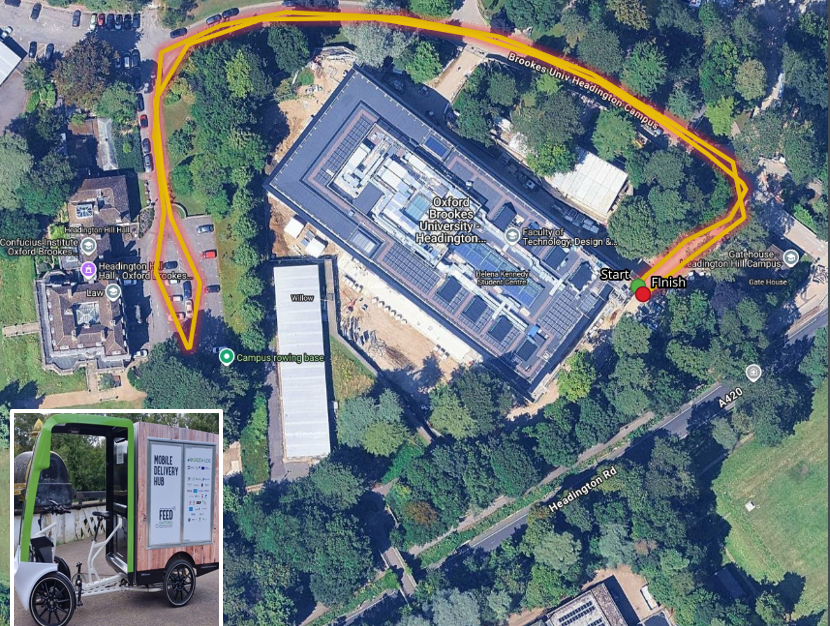}
    \caption{Online experimental route in Oxford Brookes University and GREEN-LOG teleoperated bike \cite{11654478} used for real-time evaluation.}
    \label{fig:route-teleopbike}
\end{figure}
Fig. \ref{fig:route-teleopbike} illustrates the approximately 700 m route for
the online experiment on a teleoperated delivery bike. The bike is equipped with
a ZED 2i stereo camera and Ardusimple GPS with RTK corrections via an Ardusimple
base station for ground-truth estimation. The experiments were carried out in
stereo mode under two conditions: high-contrast sunny morning and rainy
night. Each method was evaluated once per condition. In addition, each traversal
comprises three laps to ensure multiple revisits.

% !TeX root = ../main.tex

\section{Results}
\label{sec:results}
This section compares HuMem-VPR with SALAD, MegaLoc, and ORB-BoW, and evaluates HuMemSLAM against ORB-SLAM3 under varied perceptual conditions.
\subsection{VPR Accuracy and Latency}
\begin{figure}[H]
    \centering
    \includegraphics[width=\columnwidth]{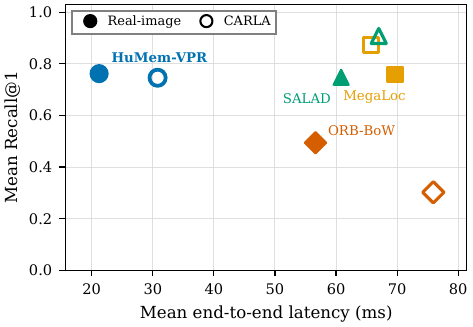}
    \caption{Standalone VPR accuracy--latency trade-off across real-image and CARLA benchmarks.}
    \label{fig:vpr-accuracy-latency-mean}
\end{figure}

Figure~\ref{fig:vpr-accuracy-latency-mean} and Table~\ref{tab:vpr-results} show
that HuMem-VPR offers a strong accuracy–latency compromise, achieving the
highest aggregate Recall@1 on the real-image benchmark while operating 2--3
times faster than the learned baselines. The advantage is most evident under the
severe KITTI B35/D80 perturbation, where HuMem-VPR reaches 0.91 Recall@1,
suggesting that its layered semantic reasoning remains discriminative when blur
and exposure degradation weaken appearance-based retrieval. However, this
advantage is not universal. Under CARLA’s strongest visibility degradation,
combined rain and fog, SALAD and MegaLoc retain higher recall. The results
therefore suggest that HuMem-VPR is overall faster and most effective when
higher-level semantic structure remains observable, while SALAD and MegaLoc are
better able to tolerate extreme scene obscuration.

\begin{table}[H]
    \caption{VPR results: (a) aggregate retrieval and latency; (b) condition-level Recall~@1.}
    \label{tab:vpr-results}
    \centering
    \footnotesize
    \begin{minipage}[t]{\columnwidth}
        \vspace{0pt}
        \centering
        \textbf{(a) Aggregate benchmark results}\par\vspace{1pt}
        \setlength{\tabcolsep}{2.6pt}
        \begin{tabular}{@{}lcccc@{}}
            \toprule
            Method & R@1 & R@5 & MRR & \shortstack{Latency\\(ms)} \\
            \midrule
            \multicolumn{5}{c}{\textbf{Real-image benchmark}} \\
            HuMem-VPR & \textbf{0.762} & 0.926 & 0.829 & \textbf{21.2} \\
            MegaLoc & 0.758 & \textbf{0.932} & \textbf{0.831} & 69.7 \\
            SALAD & 0.748 & \textbf{0.932} & 0.825 & 60.8 \\
            ORB-BoW & 0.494 & 0.694 & 0.579 & 56.6 \\
            \midrule
            \multicolumn{5}{c}{\textbf{CARLA benchmark}} \\
            HuMem-VPR & 0.746 & 0.828 & 0.778 & \textbf{30.8} \\
            MegaLoc & 0.872 & 0.962 & 0.908 & 65.7 \\
            SALAD & \textbf{0.908} & \textbf{0.988} & \textbf{0.940} & 67.0 \\
            ORB-BoW & 0.302 & 0.418 & 0.345 & 75.9 \\
            \bottomrule
        \end{tabular}
    \end{minipage}
    \par\vspace{6pt}
    \begin{minipage}[t]{\columnwidth}
        \vspace{0pt}
        \centering
        \textbf{(b) Condition-level Recall~@1}\par\vspace{1pt}
        \setlength{\tabcolsep}{2.6pt}
        \begin{tabular}{@{}lcccc@{}}
            \toprule
            \multicolumn{5}{c}{\textbf{Real-image benchmark}} \\
            Condition & HuMem-VPR & MegaLoc & SALAD & ORB-BoW \\
            \midrule
            Business Fall & 0.43 & \textbf{0.48} & 0.41 & 0.41 \\
            Garage Feb. & 0.53 & \textbf{0.59} & \textbf{0.59} & 0.47 \\
            KITTI Clean & \textbf{0.98} & \textbf{0.98} & \textbf{0.98} & 0.93 \\
            KITTI B15/D80 & 0.96 & 0.96 & \textbf{0.97} & 0.55 \\
            KITTI B35/D80 & \textbf{0.91} & 0.78 & 0.79 & 0.11 \\
            \midrule
            \multicolumn{5}{c}{\textbf{CARLA benchmark}} \\
            Condition & HuMem-VPR & MegaLoc & SALAD & ORB-BoW \\
            \midrule
            Clear Noon & \textbf{1.00} & \textbf{1.00} & \textbf{1.00} & 0.99 \\
            Deep Night & 0.93 & \textbf{1.00} & 0.99 & 0.16 \\
            Dense Fog & 0.57 & 0.77 & \textbf{0.90} & 0.11 \\
            Sunset Glare & 0.97 & \textbf{1.00} & \textbf{1.00} & 0.23 \\
            Rain + Fog & 0.26 & 0.59 & \textbf{0.65} & 0.02 \\
            \bottomrule
        \end{tabular}
    \end{minipage}
\end{table}

\begin{table*}[!t]
    \begin{minipage}{\textwidth}
    \captionof{table}{Integrated retrieval, verification, closure, and trajectory results. APE $n$ is the number of evaluable global trajectories.}
    \label{tab:integrated-slam-efficiency}
    \centering
    \scriptsize
    \setlength{\tabcolsep}{2.2pt}
    \renewcommand{\arraystretch}{1.0}
    \begin{tabular}{@{}lrrrrrrrrrrr@{}}
        \toprule
        & \multicolumn{2}{c}{Integrated R@1} & \multicolumn{3}{c}{Verification attempts} & \multicolumn{2}{c}{Applied closures} & \multicolumn{2}{c}{APE $n$} & \multicolumn{2}{c}{APE RMSE (m) $\downarrow$} \\
        \cmidrule(lr){2-3}\cmidrule(lr){4-6}\cmidrule(lr){7-8}\cmidrule(lr){9-10}\cmidrule(lr){11-12}
        Dataset & ORB & HuMem & ORB & HuMem & Red. & ORB & HuMem & ORB & HuMem & ORB & HuMem \\
        \midrule
        KITTI 06 & 0.014 & \textbf{0.617} & 27,520 & \textbf{4,292} & \textbf{84.4\%} & 53 & 58 & 60 & 59 & $1.576 \pm 1.040$ & $1.746 \pm 1.021$ \\
        KITTI 00 & 0.018 & \textbf{0.388} & 28,809 & \textbf{3,099} & \textbf{89.2\%} & 62 & 62 & 20 & 19 & $1.714 \pm 0.962$ & $1.865 \pm 1.093$ \\
        KITTI 05 & 0.042 & \textbf{0.309} & 13,991 & \textbf{1,889} & \textbf{86.5\%} & 34 & 30 & 10 & 11 & $4.947 \pm 11.659$ & $3.171 \pm 6.372$ \\
        4Seasons Garage & 0.032 & \textbf{0.324} & 30,998 & \textbf{6,218} & \textbf{79.9\%} & 124 & 81 & 29 & 29 & $3.635 \pm 1.159$ & $2.989 \pm 0.982$ \\
        4Seasons Business & 0.003 & \textbf{0.362} & 55,484 & \textbf{14,460} & \textbf{73.9\%} & 24 & 17 & 20 & 20 & $5.596 \pm 1.310$ & $5.511 \pm 1.090$ \\
        CARLA & 0.074 & \textbf{0.813} & 62,456 & \textbf{5,541} & \textbf{91.1\%} & 24 & 28 & 7 & 4 & $2.165 \pm 0.487$ & $1.871 \pm 0.325$ \\
        \bottomrule
    \end{tabular}
\end{minipage}

\end{table*}

\subsection{Integrated SLAM and Verification Efficiency}
\begin{figure}[H]
    \centering
    \includegraphics[width=\columnwidth]{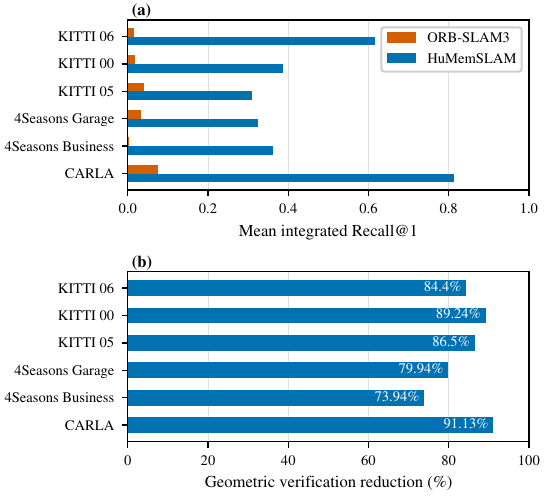}
    \caption{Integrated retrieval and verification efficiency: (a) mean Recall~@1; (b) reduction in candidates submitted to the geometric backend.}
    \label{fig:integrated-recall-verification}
\end{figure}

Figure~\ref{fig:integrated-recall-verification} and Table~\ref{tab:integrated-slam-efficiency} show that HuMemSLAM improves mean retrieval across every dataset family while reducing proposals to geometric verification by 74--91\%. More ORB-SLAM3 loop closures did not imply lower APE: despite 4 and 43 additional closures on KITTI 05 and 4Seasons Garage, respectively, its APE RMSE was higher (4.9 versus 3.2~m and 3.6 versus 3.0~m). This indicates that trajectory accuracy also depends on constraint quality and pose-graph optimisation.

\par\vspace{0pt}
\subsection{Perceptual Aliasing and System Failure Boundaries}

\begin{figure}[H]
    \centering
    \includegraphics[width=\columnwidth]{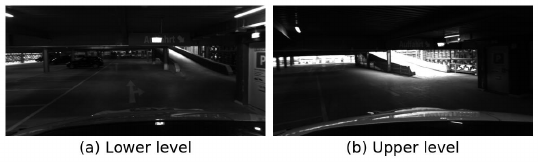}
    \caption{Visually similar observations from different garage levels.}
    \label{fig:garage-perceptual-aliasing}
\end{figure}
\begin{table}[!t]
    \caption{Closure reliability in the 4Seasons garage. Counts are accepted geometric closures; precision is the valid fraction.}
    \label{tab:garage-aliasing}
    \centering
    \footnotesize
    \setlength{\tabcolsep}{3.1pt}
    \begin{tabular}{@{}llrrrr@{}}
        \toprule
        Sequence & Method & Accepted & Valid & False & Precision \\
        \midrule
        Dec 2020 & ORB-SLAM3 & 54 & 54 & 0 & 100.0\% \\
         & HuMemSLAM & 36 & 36 & 0 & 100.0\% \\
        Feb 2021 & ORB-SLAM3 & 38 & 34 & 4 & 89.5\% \\
         & HuMemSLAM & 21 & 21 & 0 & 100.0\% \\
        May 2021 & ORB-SLAM3 & 32 & 26 & 6 & 81.3\% \\
         & HuMemSLAM & 24 & 23 & 1 & 95.8\% \\
        \midrule
        Aggregate & ORB-SLAM3 & 124 & 114 & 10 & 91.9\% \\
         & HuMemSLAM & 81 & 80 & \textbf{1} & \textbf{98.8\%} \\
        \bottomrule
    \end{tabular}
\end{table}

In the structurally repetitive and visually similar 4Seasons Garage environment
(Fig.~\ref{fig:garage-perceptual-aliasing}), ORB-SLAM3 produced 10 false loop
closures and HuMemSLAM produced 1 (Table~\ref{tab:garage-aliasing}). Although
HuMemSLAM remains susceptible to aliasing, it improves closure precision from
approximately 92\% to 99\%.

\begin{table}[!t]
    \caption{HuMemSLAM failure boundaries under severe KITTI 06 degradation.}
    \label{tab:system-failure-boundaries}
    \centering
    \footnotesize
    \setlength{\tabcolsep}{2.5pt}
    \begin{tabular}{@{}lrrrp{0.34\columnwidth}@{}}
        \toprule
        Condition & \shortstack{HuMemSLAM\\R@1} & \shortstack{Verif.\\attempts} & Closures & Observed boundary \\
        \midrule
        B35/D80 & 0.667 & 972 & 0 & Geometry rejects \\
        B35/D90 & N/A & 0 & 0 & No semantic query \\
        \bottomrule
    \end{tabular}
\end{table}

Table~\ref{tab:system-failure-boundaries} identifies two ORB-SLAM3 failure modes
limiting HuMemSLAM. At B35/D80, HuMemSLAM retrieved candidates, but severe
degradation prevented geometric verification. At B35/D90, the frontend generated
no viable keyframes, so HuMemSLAM remained idle.

\subsection{Ablation Study}
\begin{table}[H]
    \caption{HuMem-VPR ablation. Full denotes SCAL+OSRL+OGTL.}
    \label{tab:ablation}
    \centering
    \footnotesize
    \setlength{\tabcolsep}{2.5pt}
    \begin{tabular}{@{}lcccc@{}}
        \toprule
        \multicolumn{5}{c}{\textbf{Aggregate performance}} \\
        Metric & SCAL & \shortstack{SCAL+\\OSRL} & \shortstack{SCAL+\\OGTL} & Full \\
        \midrule
            Mean R@1 $\uparrow$ & 0.711 & 0.746 & 0.751 & \textbf{0.757} \\
            Mean R@5 $\uparrow$ & 0.842 & 0.878 & 0.870 & 0.877 \\
            Mean MRR $\uparrow$ & 0.797 & 0.804 & 0.798 & 0.803 \\
            Mean latency (ms) $\downarrow$ & 13.7 & 21.3 & 24.3 & 25.1 \\
            Mean p95 (ms) $\downarrow$ & 15.0 & 26.4 & 31.3 & 32.9 \\
        \midrule
        \multicolumn{5}{c}{\textbf{Condition-level Recall~@1}} \\
        Condition & SCAL & \shortstack{SCAL+\\OSRL} & \shortstack{SCAL+\\OGTL} & Full \\
        \midrule
            Business Fall & 0.30 & 0.35 & \textbf{0.43} & \textbf{0.43} \\
            Garage Feb. & 0.45 & 0.53 & \textbf{0.55} & \textbf{0.55} \\
            KITTI Clean & \textbf{0.98} & \textbf{0.98} & \textbf{0.98} & \textbf{0.98} \\
            KITTI B15/D80 & \textbf{0.97} & 0.96 & \textbf{0.97} & \textbf{0.97} \\
            KITTI B35/D80 & 0.89 & \textbf{0.91} & \textbf{0.91} & \textbf{0.91} \\
            CARLA Clear Noon & \textbf{1.00} & \textbf{1.00} & \textbf{1.00} & \textbf{1.00} \\
            CARLA Deep Night & 0.92 & \textbf{0.93} & \textbf{0.93} & \textbf{0.93} \\
            CARLA Dense Fog & 0.45 & \textbf{0.57} & 0.54 & \textbf{0.57} \\
            CARLA Rain + Fog & 0.20 & \textbf{0.26} & 0.23 & \textbf{0.26} \\
            CARLA Sunset Glare & 0.95 & \textbf{0.97} & \textbf{0.97} & \textbf{0.97} \\
        \bottomrule
    \end{tabular}
\end{table}

Table~\ref{tab:ablation} shows that SCAL is a strong low-latency baseline.
SCAL+OGTL gives higher mean Recall~@1, whereas SCAL+OSRL gives higher mean
Recall~@5 and MRR at lower latency. Full achieves the best mean Recall~@1
(0.757). Gains vary by condition: the best configuration improves Business Fall,
Garage Feb., and CARLA Dense Fog by 13, 10, and 12 Recall~@1 points over SCAL.
Thus, OSRL and OGTL provide complementary, condition-dependent evidence for the
global SCAL hypothesis.

\subsection{Real-World Online Evaluation}

Fig. \ref{fig:realworld-evaluation} shows that HuMemSLAM achieved higher
Recall@1 than ORB-SLAM3, most notably under rainy-night condition, whilst
reducing geometric verification attempts by approximately 63--65\%
across both traversals.

\begingroup
\setlength{\intextsep}{6pt}
\begin{figure}[H]
    \centering
    \includegraphics[width=\columnwidth]{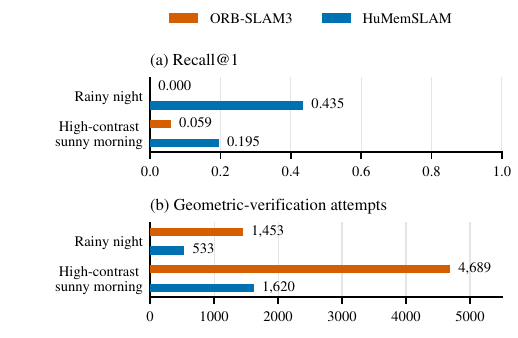}
    \caption{Real-world online evaluation: (a) Recall~@1 and (b) geometric-verification attempts on the two campus runs.}
    \label{fig:realworld-evaluation}
\end{figure}
\endgroup

\FloatBarrier

% !TeX root = ../main.tex

\section{Conclusion}
\label{sec:conclusion}

We presented HuMem-VPR, a novel human-inspired place-recognition method,
and HuMemSLAM, its integration with ORB-SLAM3. Across the evaluated datasets,
HuMemSLAM substantially improved integrated Recall~@1 while reducing proposals
to the geometric backend by approximately 74--91\%. In the perceptually
aliased 4Seasons Garage, false loop closures were reduced from 10 to 1.
HuMem-VPR also achieved the highest aggregate retrieval accuracy on the
real-image benchmark and competitive CARLA performance while operating
approximately two to three times faster than the evaluated learned VPR
baselines. Severe-degradation experiments nevertheless exposed a dependency on
the underlying geometric pipeline: successful semantic retrieval cannot yield a
SLAM correction when viable keyframes or geometric correspondences are absent.

These results suggest that low-latency semantic retrieval can support more
responsive visual SLAM while reducing unnecessary geometric computation, an
increasingly important capability for autonomous vehicles operating at higher
speeds or in rapidly changing environments. Future work will investigate
learned local features to strengthen geometric verification under severe
appearance degradation, alongside INT8 quantisation and further hardware-aware
optimisation to reduce inference latency and memory requirements. Together,
these directions aim towards robust semantic localisation under progressively
tighter computational and temporal constraints.

\bibliographystyle{IEEEtran}
\bibliography{bib/references}

\end{document}